\documentclass[sigconf]{acmart}

\usepackage[small,compact]{titlesec}
\usepackage{multicol}
\usepackage{siunitx}
\usepackage{pdflscape}
\usepackage{graphicx}
\usepackage{tabularray}
\usepackage{xcolor}
\usepackage{colortbl}
\usepackage{lipsum}

\setcopyright{acmcopyright}

\usepackage{xcolor}

\usepackage[linesnumbered,ruled,vlined]{algorithm2e}

\SetKwComment{Comment}{\color{green!50!black}// }{}

\SetKwProg{Function}{function}{}{}

\copyrightyear{2023}
\acmYear{2023}
\setcopyright{rightsretained}
\acmConference[IDEAS 2023]{International Database Engineered Applications Symposium Conference}{May 5--7, 2023}{Heraklion, Crete, Cyprus} 
\acmBooktitle{International Database Engineered Applications Symposium Conference (IDEAS 2023), May 5--7, 2023, Heraklion, Crete, Cyprus}
\acmDOI{10.1145/3589462.3589498} 
\acmISBN{979-8-4007-0744-5/23/05}

\begin{document}

\title{AdapterEM: Pre-trained Language Model Adaptation for Generalized Entity Matching using Adapter-tuning}
  
\renewcommand{\shorttitle}{AdapterEM: PrLM Adaptation for Generalized EM using Adapter-tuning}

\author{John Bosco MUGENI}
\authornote{first author}
\orcid{0000-0002-2464-413X}
\affiliation{%
  \institution{University of Tsukuba}
  \streetaddress{1-1-1 Tennodai,}
  \city{Tsukuba} 
  \country{Japan}}
  \postcode{305-8577}

\email{bosco@kde.cs.tsukuba.ac.jp}

\author{Steven LYNDEN}
\affiliation{%
  \institution{National Institute of Advanced Industrial Science \& Technology (AIST)}
  \streetaddress{}
  \city{Tokyo} 
  \country{Japan}}
\email{steven.lynden@aist.go.jp}

\author{Toshiyuki AMAGASA}
\affiliation{%
  \institution{University of Tsukuba}
  \streetaddress{1-1-1 Tennodai,}
  \city{Tsukuba} 
  \country{Japan}}
  \postcode{}

\email{amagasa@cs.tsukuba.ac.jp}

\author{Akiyoshi MATONO}
\affiliation{%
  \institution{National Institute of Advanced Industrial Science \& Technology (AIST)}
  \streetaddress{}
  \city{Tokyo} 
  \country{Japan}}
  \state{} 
  \postcode{}

\email{a.matono@aist.go.jp}

\renewcommand{\shortauthors}{Mugeni et al.}

\begin{abstract}
Entity Matching (EM) involves identifying different data representations referring to the same entity from multiple data sources and is typically formulated as a binary classification problem. It is a challenging problem in data integration due to the heterogeneity of data representations. State-of-the-art solutions have adopted NLP techniques based on pre-trained language models (PrLMs) via the fine-tuning paradigm, however, sequential fine-tuning of overparameterized PrLMs can lead to catastrophic forgetting, especially in low-resource scenarios. In this study, we propose a parameter-efficient paradigm for fine-tuning PrLMs based on adapters, small neural networks encapsulated between layers of a PrLM, by optimizing only the adapter and classifier weights while the PrLMs parameters are frozen. Adapter-based methods have been successfully applied to multilingual speech problems achieving promising results, however, the effectiveness of these methods when applied to EM is not yet well understood, particularly for generalized EM with heterogeneous data. Furthermore, we explore using (i) pre-trained adapters and (ii) invertible adapters to capture token-level language representations and demonstrate their benefits for transfer learning on the generalized EM benchmark. Our results show that our solution achieves comparable or superior performance to full-scale PrLM fine-tuning and prompt-tuning baselines while utilizing a significantly smaller computational footprint $\approx 13\%$ of the PrLM parameters.

\end{abstract}

%
%



\maketitle

\section{Introduction}
\label{section:machmp}
\begin{figure*}[h!]
    \centering
    \includegraphics[width=15cm]{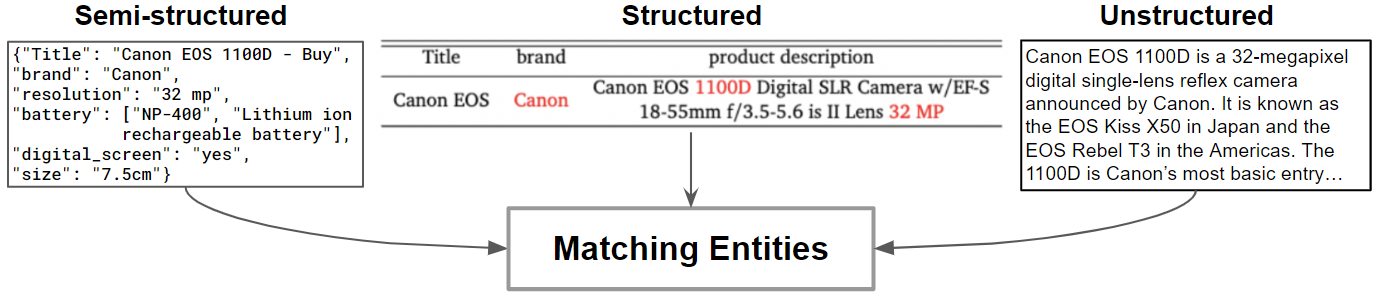}
    \caption{An example of Generalized Entity Matching.}
    \label{fig: gem}
\end{figure*}

Entity matching (EM) is a long-established problem in the database community which aims to identify records as matching or non-matching entries among heterogeneous data sources~\cite{a1}. Until recently, EM approaches have focused on more traditional settings that are simplistic and not consistent with the nature of EM in the real world~\cite {a2}. Such approaches are often evaluated using benchmarks constructed with various simplifying assumptions which do not hold in practical, real-world scenarios and can exaggerate the performance of the developed techniques~\cite{a2}. For example, traditional EM assumes only structured tables, an assumption that is not representative of real-world applications where data might exist in heterogeneous formats, e.g., JSON, text, etc. Furthermore, traditional EM builds on the notion of schema identicality, which also may not conform to the characteristics of real-world EM. For these reasons, a recently published research benchmark, Machamp~\cite{a3}, was constructed with the aim of encapsulating such real-world scenarios. An illustration of the key features of the Machamp benchmark is shown by the example entities from an E-commerce context in Figure \ref{fig: gem}. Due to various reasons in 
real-world scenarios, sources might provide information as plain text (right), structured (middle) data, for example, relational tables, or semi-structured data (left) such as JSON objects or XML which may add more complexity, e.g. nested objects. This also entails that unifying the schema of such data would be a non-trivial process, for example, matching schemaless textual instances with relational ones. Additionally, Machamp also relaxes the classic EM criteria to a more general one, such as two entities being relevant to each other rather than identical. The problem of matching such entities is referred to as \textit{Generalized Entity Matching} (GEM).  

Meanwhile, in the NLP domain, extremely deep neural networks called Transformers~\cite{a4}, pre-trained on large text corpora, have dominated the domain and its sub-disciplines, taking over from conventional methods such as support vector machines (SVM's) and recurrent neural networks. However, due to the overparameterization of pre-trained language models (PrLMs), they can suffer from catastrophic forgetting, especially in low-resource scenarios \cite{task_to_task,a10}. This happens when the PrLM forgets its pre-training knowledge during downstream task adaptation \cite{a10}. Moreover, sequentially fine-tuning a PrLM for each new task can introduce several drawbacks. 
\begin{figure}[h!]
    \centering
    \includegraphics[width=8cm]{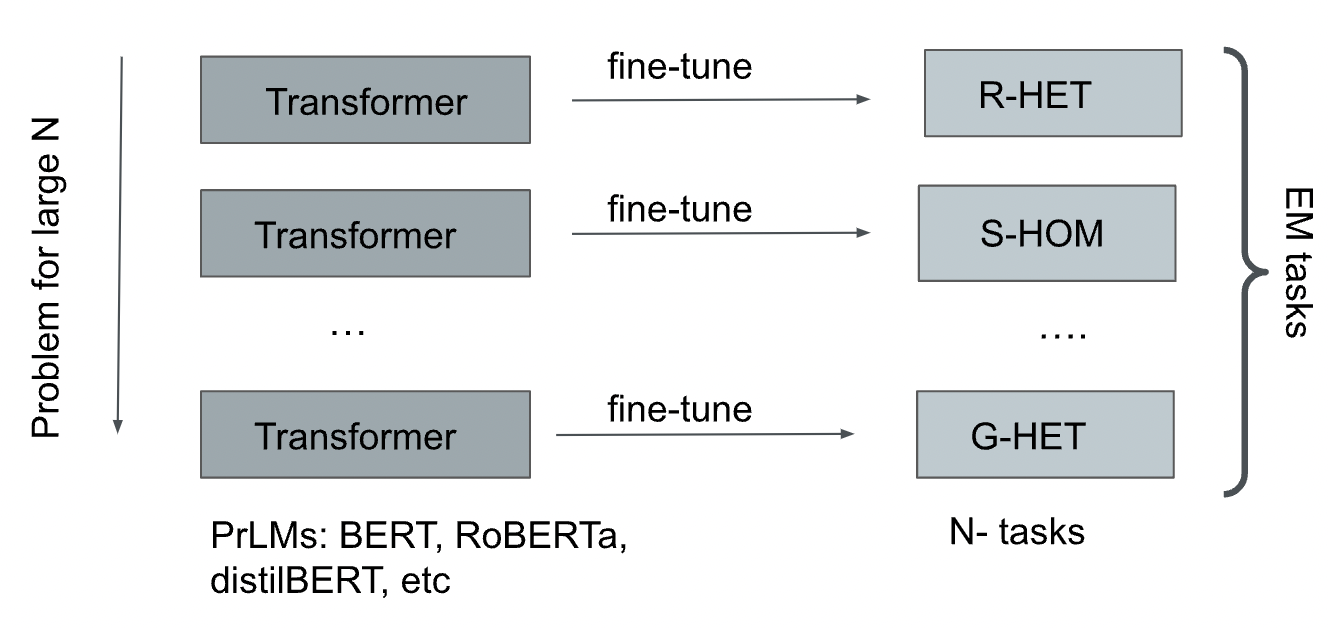}
    \caption{Full-scale PrLM fine-tuning: this results in additional storage space for each model checkpoint (e.g., 1.3GB x N tasks).}
    \label{fig: full-fine-tune}
\end{figure}
For example, fine-tuning a PrLM for each of our $N$ tasks will require a new checkpoint for each task (Figure~\ref{fig: full-fine-tune}). This incurs additional storage costs as $N$ grows. As an example of this, the JoinBERT \cite{a6}  checkpoint\footnote{\url{http://data.dws.informatik.uni-mannheim.de/largescaleproductcorpus/data/v2/repo-download/jointbert.zip}} on the web data commons (WDC) Computers xlarge dataset \cite{a13} amasses $\sim$ 1.3GB of disk storage, which potentially makes storage and sharing model checkpoints prohibitive for N- tasks.

Recently, a new paradigm of efficiently fine-tuning language models, which can match the performance of full-scale PrLM fine-tuning, has emerged called adapter-tuning~\cite{pfeiffer2020AdapterHub}.
\begin{figure}[t!]
    \centering
    \includegraphics[width=8cm]{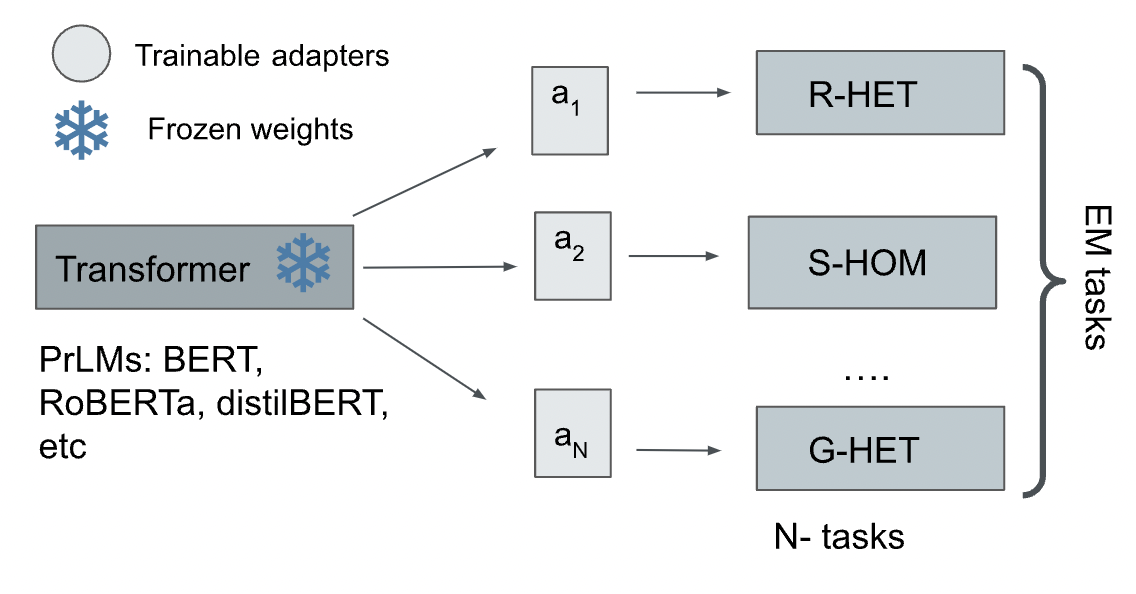}
    \caption{Adapter-based approach: this results in efficient storage space used for each adapter checkpoint (~ $7MB \times N$ tasks).}
    \label{fig: adapters}
\end{figure}
In lieu of modifying the parameters $\Phi$ of the PrLM, adapter-tuning adds a new set of parameters $\phi$ called adapters between Transformer blocks ($\sim$ 1\% of $\Phi$). During fine-tuning, we update $\phi$ while keeping $\Phi$ fixed. Adapters are stitched into a PrLM as task-specific or language-specific: task-specific adapters are optimized on the target task whereas language-specific adapters are optimized via masked language modeling (MLM) on target task data without labels~\cite{a18}. To this end, there are several benefits to using adapter-tuning. Firstly, we keep one generalist model and train new adapters with few parameters $\phi$ for each of our N tasks (Figure \ref{fig: adapters}), with each adapter typically occupying approximately 7MB of disk storage, making them easily shareable for inference purposes compared with the much larger size of a PrLM. Since adapter parameters $\phi$ $\ll$ $\Phi$ of the PrLM, they can significantly reduce training times and memory usage, which is beneficial when implementing EM and an important consideration in general, demonstrated by the adoption of computational and carbon footprint monitoring tools in popular cloud computing platforms such as AWS~\footnote{\url{https://aws.amazon.com/}}.

Furthermore, adapters are composable, i.e., we can stack, fuse or mix different adapters to leverage their combined knowledge. In view of this, we hypothesize that adapter-tuning can also minimize catastrophic forgetting (i.e., the knowledge gap between the objective forms of pre-training and full-scale PrLM fine-tuning), with far less computation.

In this work, we study GEM using the recently introduced adapter-tuning~\cite{pfeiffer2020AdapterHub} paradigm. To the extent of our knowledge, adapter-tuning remains unexplored for both EM and GEM in general. Thus, we introduce AdapterEM, a supervised system for GEM that leverages adapter-tuning as the underlying mechanism. We conduct experiments for both low-resource (i.e., the Machamp training set has fewer examples) and sufficient-resource (i.e., the entire Machamp dataset) settings, and specifically, our contributions are as follows: 

\begin{enumerate}
    \item We introduce a system that leverages transfer learning via efficient adapter-tuning, which can sometimes match the classification accuracy of the state-of-the-art prompt tuning-based PromptEM~\cite{a12}, using much less memory overhead. 
    \item We investigate the impact of composing task and language-specific adapters for the GEM problem in addition to using task adapters.
    \item Our results empirically demonstrate that our system can alleviate catastrophic forgetting on 5 out of 7 tasks in the Machamp benchmark for both low and sufficient-resource settings compared to full-scale PrLM fine-tuning. Additionally, we also consider a benchmark from the geospatial domain for which we minimize catastrophic forgetting. 
    \item We also make our pre-trained adapters available to the community to help accelerate research into the future use of adapters for the EM domain. Our code is publicly available on GitHub\footnote{\url{https://github.com/boscoj2008/AdapterEM}}.
\end{enumerate}

The remainder of this paper is organized as follows. Section~\ref{section:Background} presents the background area, highlights some of the major research problems, and discusses related work. Section~\ref{fig: adapters} introduces the proposed approach. Section~\ref{section: Problem Setting} introduces the training of adapters, followed by Section~\ref{section: Experiments} which details the evaluation benchmark and experimental setup. The results are presented and discussed in Section~\ref{section: Results}, and lastly, Section~\ref{section: Conclusion} concludes this work.

\section{Background}
\label{section:Background}
\subsection{Full-scale PrLM Fine-tuning For EM/GEM}

Recently, Transformers have become the cornerstone of text processing in NLP. This is owing to their highly contextual embeddings which can capture polysemantic word relationships~\cite{a5}. Typically, they ingest encoded text pairs, extract word embedding vectors and run them through multiple self-attention mechanisms embedded within and across a stack of N~ identical Transformer blocks (e.g., 12 blocks for  BERT base). The final block produces these highly contextualized embeddings summarized into a special \textit{[CLS]} token which is inputted to a shallow network to learn class probabilities. At training time, the pre-trained network parameters $\Phi$ are fine-tuned jointly with the shallow network to learn the new task, a process we shall refer to as full-scale PrLM fine-tuning. In EM, this strategy has been widely adopted. For example, Ditto~\cite{a5}, can leverage additional modules such as data augmentation, knowledge injection, and text summarization. JointBERT~\cite{a6} can leverage either single or dual objective training depending on whether the dataset contains auxiliary identification numbers or not. Rotom \cite{a8} which utilizes a meta-learning framework with a suite of data augmentation strategies for low-resource settings. DADER~\cite{a9} which defines a solution space consisting of feature extraction, alignment, and matching modules via domain adaptation. Additionally,  Brunner et al.~\cite{a7} which studies different Transformer methods using the traditional EM benchmarks described in Magellan~\cite{a17}. However, all the above methods are prone to catastrophic forgetting (with the exception of DADER which possibly mitigates this through domain adaptation) and consist of an expensive parameter update.

\subsection{Prompt-tuning For GEM}
Parameter-efficient methods for fine-tuning PrLMs have become attractive for mitigating the above issues and are a hotly contested research direction evidenced by a large volume of recently published work. A seminal work focused on investigating parameter efficient tuning for the GEM problem is PromptEM \cite{a11}, which leverages the underlying mechanism of prompt-tuning, achieving a new state-of-the-art for GEM. At the core, prompt-tuning re-purposes the original parameters $\Phi$ of the PrLM for different tasks while holding them fixed. The MLM objective then serves as a query via designated templates called prompts. These prompts involve additional tokens prepended to a serialized input $X$ of the PrLM via a prompt template $T(\cdot)$ which is conditioned to produce the correct class label for Y, $P_{\Phi} (Y| T(x))$ \cite{a12}.

However, it is important to note that self-attention scales quadratically w.r.t the input sequence in Transformers \cite{a4}, i.e., $(O(n^2 \cdot d))$, where $n$ is the sequence length and $d$ is the representation dimension. Prompt-tuning inadvertently adds to sequence length which can increase computational costs as well as potentially lengthen training times, especially across language model scales. 

\subsection{Prefix-tuning for EM}
Another paradigm that involves efficient tuning of PrLMs has gained attention within the context of EM called prefix-tuning~\cite{prefix-tuning}. Instead of using manually engineered prompt templates, whose design process is non-trivial, prefix-tuning prepends virtual tokens called prefixes to the input sequence of the PrLM and learns them as continuous prompts. Similar to prompt-tuning, prefix-tuning allows us to maintain a generalist model for all tasks without modifying the PrLM parameters. However, prefix-tuning performs sub-optimally compared to full-scale PrLM fine-tuning~\cite{prefix-paper} on traditional EM benchmarks. It should also be noted that prior works~\cite{prefix-paper} have not investigated prefix-tuning for GEM. 

\section{Combining Transformers with Adapters}
\label{section:Adapters}
\subsection{Adapter Architecture}

\begin{figure}[h!]
    \centering
    \includegraphics[width=5cm]{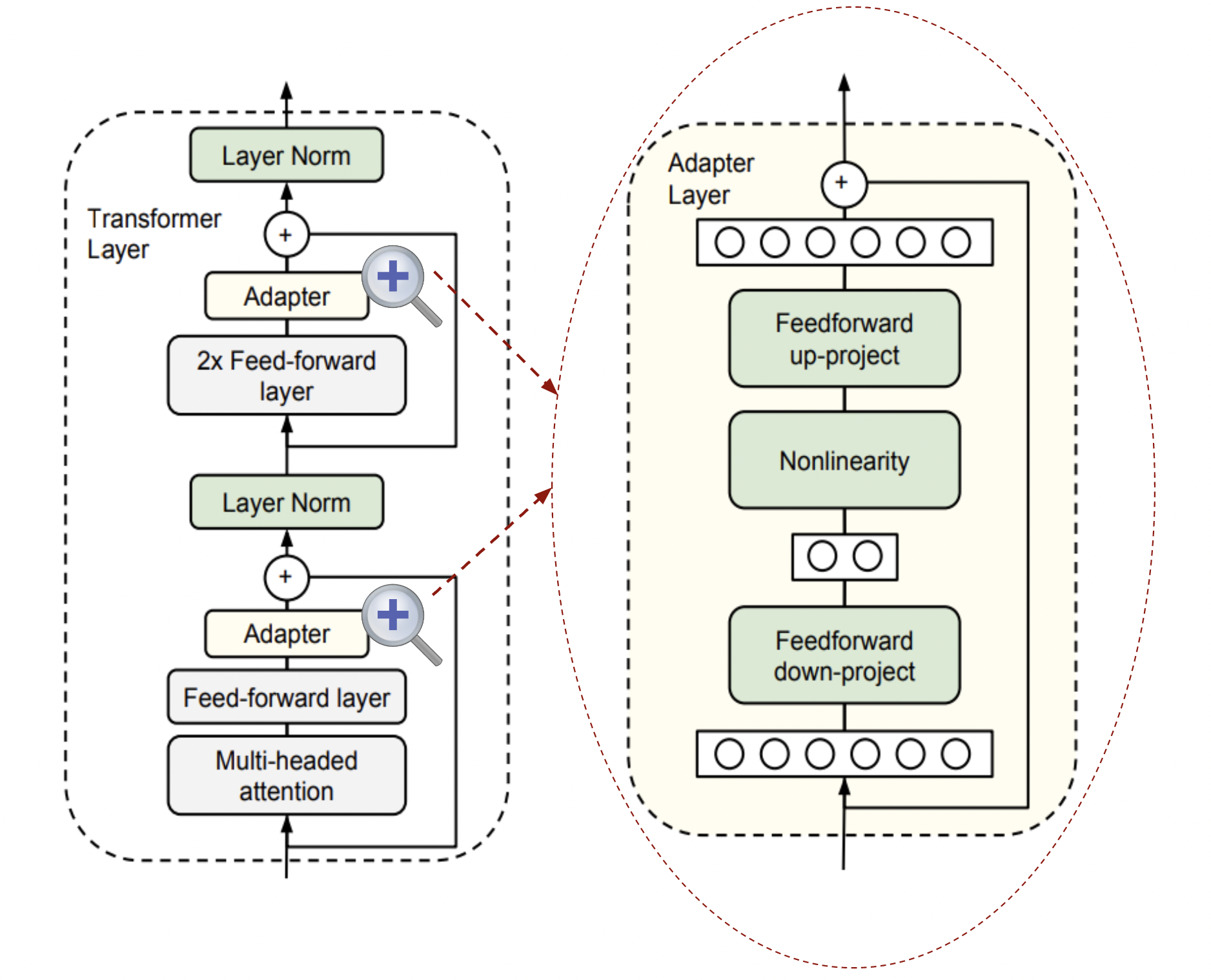}
    \caption{The Houlsby adapter proposed in \cite{houlsby2019parameterefficient}.}
    \label{fig: houlsby}
\end{figure}

Adapters were first applied to the computer vision domain to learn domain-specific representations~\cite{a19}. Subsequently, adapters have gained remarkable attention within the context of NLP research for efficient PrLM adaptation and transfer learning. Houlsby et al.~\cite{houlsby2019parameterefficient} introduce adapters as small bottleneck layers added to each Transformer layer -- utilizing two down and up-projections (Figure \ref{fig: houlsby}) -- and later Pfeiffer et al.~\cite{a18} propose a variant with a single down and up-projection. In our preliminary runs, we find the Houlsby architecture yields favorably for our tasks. Thus we utilize it as the underlying setting for our adapters.

The Transformer layer consists of two sub-layers: an attention layer and a feedforward layer. Immediately following these layers is a projection that restores the original input dimension. Typically, a residual connection is placed across the sub-layers with the output being fed to the layer normalization (Figure \ref{fig: houlsby}). Adapters are inserted serially after these sub-layers and the aforementioned projection, however, before the residual connection. In order to regulate the number of parameters, the Houlsby adapter proposes a bottleneck architecture. This architecture consists of down and up projection operations. The down projection reduces the original input dimension $d$ to a small dimension $m$, the bottleneck size, applies a non-linearity, and then an up projection to restore the original input dimension $d$. Thus, the number of parameters introduced at each layer including biases becomes $2md + d + m$. Essentially, we choose $m \ll d$, resulting in the usage of less than 13\% of PrLM $\Phi$~\cite{houlsby2019parameterefficient}. It should also be noted that the Houlsby architecture consists of its own residual connection to initialize module parameters to an approximate identity function if they are initialized near zero. 

\subsection{Adapter Configurations}
\begin{figure}[h!]
    \centering
    \includegraphics[width=8.5cm]{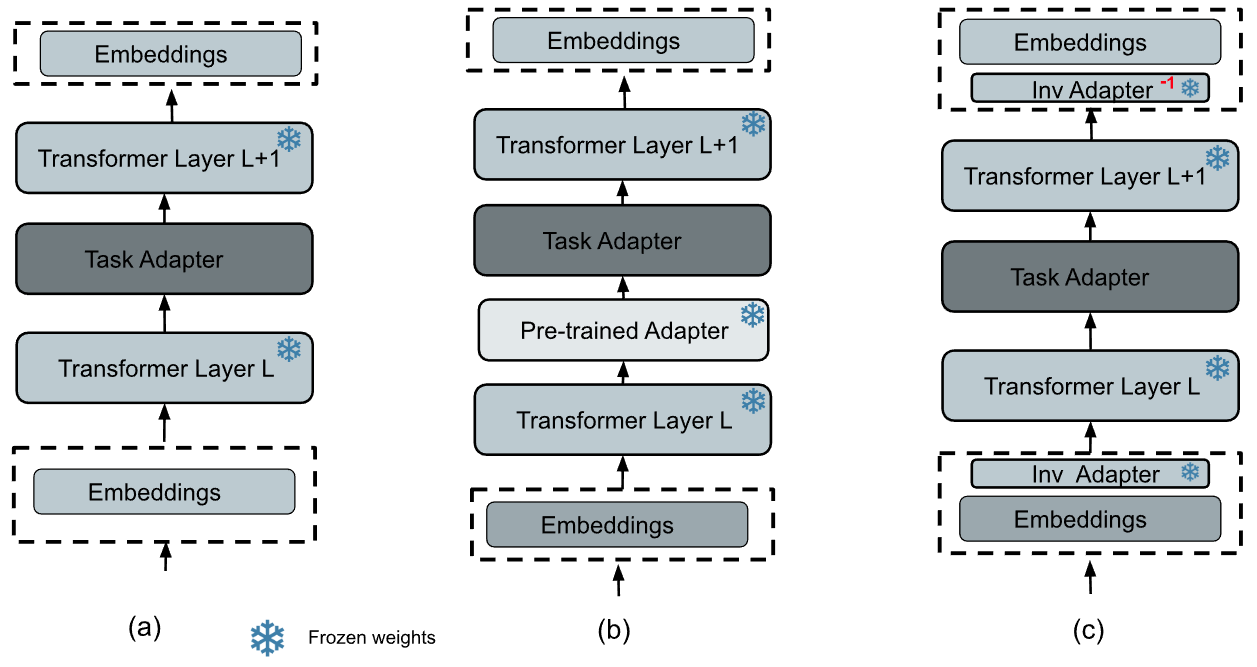}
    \caption{Adapter setup; (a) task adapter only, (b) task adapter stacked on top of language adapter, (c) invertible adapter with task adapter.}
    \label{fig: setup}

\end{figure}

There are two ways to utilize and combine adapters for downstream tuning; parallel or vertical stacking. We have adopted vertical stacking as the building block of our experiments (see Figure \ref{fig: setup}). From this, we devise three strategies; (i) using a task adapter in each Transformer block (Figure \ref{fig: setup}(a)), (ii) fusing a pre-trained adapter with a task adapter in each Transformer block (Figure \ref{fig: setup}(b)) and (iii) fusing an invertible language adapter with a task adapter in each Transformer block (Figure \ref{fig: setup}(c)). It should be noted that in each setting, the task adapter is randomly initialized.

Specifically, a pre-trained adapter contains parameters somewhat relevant to our downstream task, e.g., an adapter trained on the Stanford Natural Language Inference (SNLI) corpus~\cite{snli:emnlp2015}. Several pre-trained adapters are publicly available on AdapterHub\footnote{\url{https://adapterhub.ml/explore/}}, an adapter-sharing platform. Thus, for the Transformer block, we can easily stitch in a pre-trained adapter, stack a new (randomly initialized) task adapter atop and update its parameters during fine-tuning. This configuration is illustrated in Figure \ref{fig: setup}(b) and is in stack contrast to using a task adapter only (Figure \ref{fig: setup}(a)).

\

On the other hand,  the invertible language adapter captures token-level language-specific representations via MLM training. The goal is to adapt it to domain-specific terminology unseen during original pre-training (also called task-adaptive pre-training (TAPT)). In terms of architecture, the invertible adapter is identical to the adapter presented in Section \ref{section:Adapters} (Figure \ref{fig: houlsby}); however, it involves an invertible adapter layer after the input embedding layer (i.e., before the first Transformer layer) and in the inverse direction at the output (i.e., after the last Transformer layer, see Figure \ref{fig: setup}(c)). Finally, we load the trained invertible adapter before a task adapter for downstream model tuning.

\section{Training Adapters For GEM}
\label{section: Problem Setting}
We will now describe the serialization method adopted for the Machamp benchmark and a brief overview of the task definition for GEM.

\subsection{Serialization}
Existing techniques developed for EM are generally optimized for structured data with homogeneous schema only. Since GEM contains data of different data formats (i.e., structured, semi-structured, or unstructured), we adopt a serialization process, proposed in prior works~\cite{a3,a11}, to convert these data types into token sequences for the PrLM, summarized in Algorithm 1.
\vspace{0.05in}

\begin{figure}[h!]
    \centering
    \includegraphics[width=8.5cm]{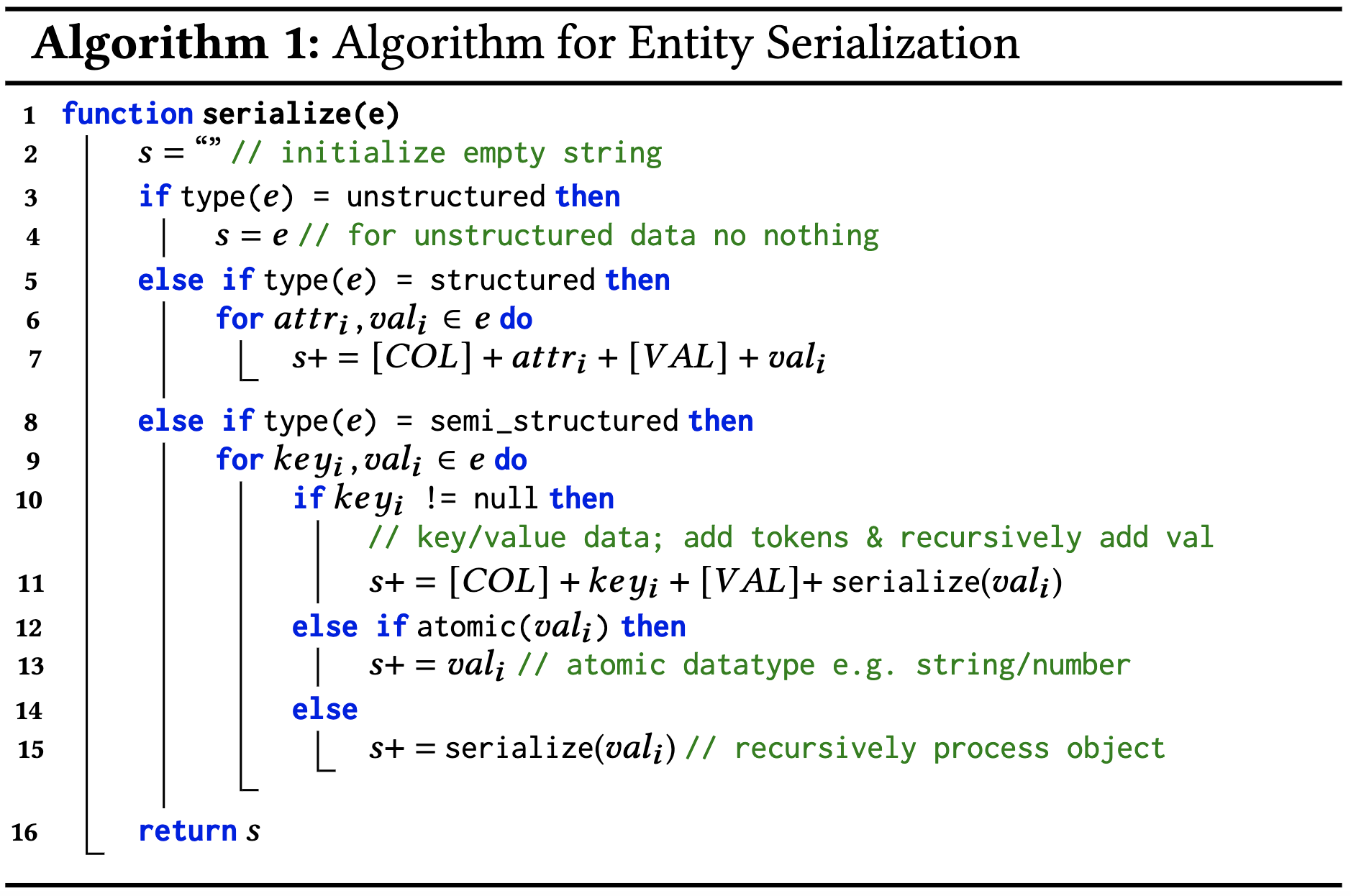}
    \label{fig:serialize}

\end{figure}

The algorithm adds special tokens to encode structured data, for example, the string

\vspace{0.1in}

``\small{\emph{[\textbf{COL}] Title [\textbf{VAL}]Canon EOS 1100D [\textbf{COL}] brand [\textbf{VAL}]Canon[\textbf{COL}] product description [\textbf{VAL}] Digital SLR Camera w/EF-S 18-55mm f/3.5-5.6 is II Lens 32MP}}'' 

\vspace{0.1in}

will be produced for the structured data in Figure~\ref{fig: gem}. For the semi-structured data in the same figure, the output is 

\vspace{0.1in}

``\small{\emph{[\textbf{COL}] Title [\textbf{VAL}] Canon EOS 1100D - Buy [\textbf{COL}] brand [\textbf{VAL}] Canon [\textbf{COL}] battery [\textbf{VAL}] NP-400 Lithium, ion rechargeable battery [\textbf{COL}] digital\_screen [\textbf{VAL}] yes [\textbf{COL}] size [\textbf{VAL}] 7.5cm}}''. 

\vspace{0.1in}

Note that the \texttt{+} (string concatenation) operator in the serialization algorithm adds commas in between list items in semi-structured data.

\subsection{Task definition:} Let $X=\{x_1 \dots x_n \}$ be a set of candidate pairs and $Y=\{y_1 \dots y_n \}$ a label set. Given the above operations, an entity pair x := $\langle$$e$, $e'$$\rangle$ is serialized and concatenated by the usual  [\textbf{SEP}] token. Also, the  [\textbf{CLS}] token is prepended at the beginning to denote the start of a sequence, i.e., 

\[ x := [\textbf{CLS}] serialize(e) [\textbf{SEP}] serialize(e') [\textbf{SEP}] \]

\vspace{0.05in}

The PrLM uses the [\textbf{CLS}] pretext to encode the sequence pair into an n-dimensional vector. Next, we describe how this vector is further utilized for GEM. Since the PrLM parameters are composed of $\Phi$ (frozen set of weights) and $\phi$ (the adapter parameters), we only update $\phi$ according to a loss function $\ell$.   Let $f(\cdot;\{\Phi, \phi\})$ denote a projection $f$ associated with the parameter $\phi$ from the input space (our vector) to an output space. Since we are solving a classification problem, the output space corresponds to a k-standard simplex, where k is equal to the number of classes. Conceptually, we define the standard simplex as a convex closure whose dimension is $k$, i.e., 

\[ \{\mathbf{z} \in \mathbf{R}^k: z_0+ \dots +z_{k-1} = 1,  z \geq 0 \; \mbox{for} \; i = 0, \dots, k-1 \} \]

\vspace{0.05in}

\noindent
where $\textbf{z}$ is a set of defining points called vertices. We then denote the PrLMs final logits as $\textbf{z} := g(\cdot;\{\Phi, \phi\})$ s.t $f(\cdot;\{\Phi, \phi\}) = \sigma (g(\cdot;\{\Phi, \phi\}))$ where $\sigma(\cdot)$ is the softmax function. The training samples are denoted as $ \{ x_i, y_i \} |_{i=1}^n $. For our case, we compute the training loss for $f(\cdot;\{\Phi, \phi\})$ as follows:


\begin{equation}
    \phi^* \leftarrow \mathop{\arg \min}_{\phi} \ell(f(\cdot;\{\Phi, \phi\}), y_i)
\end{equation}

\noindent for the training samples ($X$, $Y$) where $\ell(\cdot ; \cdot)$ is the cross-entropy loss function:

\begin{equation}
   CELoss = -  \sum\limits_{i=1} y_i \log(f(\cdot;\{\Phi, \phi\})) 
\end{equation}

\subsection{Training}
In this section, we describe the process of training the task and invertible adapters.
\subsubsection{Adapter Training:}
 Adapter-tuning is implemented in the adapter-transformers~\cite{pfeiffer2020AdapterHub} library, where two seminal adjustments are proposed: (i) loading the adapter and (ii) activating the adapter. Thus, we proceed as follows; after adding the adapter(s), the PrLM's parameters are frozen. If a pre-trained adapter is used, its pre-trained weights are also frozen. It is also possible to unfreeze these weights, which can give the PrLM more expressive power; however, unfreezing these weights is beyond the scope of this work. Next, the PrLM's remnant parameters owing to the (randomly initialized) task adapter and the prediction head of the PrLM are  fine-tuned on the downstream task (i.e., activated) and updated during backpropagation.

\subsubsection{MLM Training:}
As mentioned before, invertible adapters are trained via the MLM objective. For this, the task data without labels is leveraged as a training signal. It should be noted that during MLM training, the PrLM parameters $\Phi$ are also frozen while the adapter parameters $\phi_{inv}$ are updated. Furthermore, we are inspired by the works of~\cite{MLM_train}, whereby the masking probability $p$ is set slightly higher for MLM. This allows us to corrupt a larger proportion of tokens. Thus, the invertible adapter would conceivably learn better representations under these settings by making more predictions. This is designed to help understand the impact of masking probability on the invertible adapters. 

\begin{table*}[t!]
  \begin{center}
    \caption{Dataset statistics for low-resource settings$^\dag$.}
    \label{tab:dat_stats}
    \scalebox{1.1}{
    \begin{tabular}{|l|l|l|c|c|c|} 
      \hline
      \textbf{Dataset} &  \textbf{Domain} &\textbf{\#Tuples (L -  R)} & \textbf{\#Labeled} & \textbf{Train \% rate}& \textbf{Train set (P\textbackslash N)} \\

      \hline

      R-HET & restaurant & 534 - 332 & 567 & 10\% & 7 \textbackslash 50\\
      S-HOM & citation & 2,616 - 64,263 & 17,223 & 5\% & 160\textbackslash 701 \\
      S-HET & book & 22,133 - 23,264 & 1,240 & 10\% &  47 \textbackslash 77 \\
      S-REL & movie & 29,180 - 32,823 & 1,309 & 10\% & 54\textbackslash 76\\  
      S-TEX-w & product & 9,234 - 9,234 & 5,540 & 10\% & 64 \textbackslash 490\\
      S-TEX-c & product & 20,897 - 20,897 & 12,538& 5\%& 89 \textbackslash 538 \\
      R-TEX & citation & 2,616 - 2,295  & 7,417 & 10\%& 133 \textbackslash 608\\
      G-HET & geo-spatial &2,469 - 2,788  & 2,500 & 10\% & 73 \textbackslash 177\\

      \hline
 
    

    \multicolumn{6}{c}{%
    \begin{minipage}{9cm}%
      \small $^\dag$ Note: For sufficient-resource setting, \% rate is set to 100\%, i.e., all the available training data. P denotes positive (matching) pairs and N denotes negative (non-matching) pairs. L and R denote the left and right tuples of respective datasets.
     \end{minipage}%
      }\\

   \end{tabular}}  
  \end{center}

\end{table*}

\section{Experiments}
\label{section: Experiments}

Machamp consists of datasets derived from Magellan~\cite{a17}, DeepMatcher~\cite{deepmatcher}, and the WDC product~\cite{a13} datasets to construct GEM-relevant subtasks. It encapsulates the challenges discussed in Section~\ref{section:machmp} making it the most complete and extensive benchmark to date for GEM that provides labeled data between various data representations, where each dataset consists of a left and right set of entities represented as follows: R-HET: matching between structured tables with heterogeneous schema; S-HOM: matching between semi-structured data with homogeneous schema; S-HET: matching between semi-structured data with heterogeneous schema; S-REL: matching semi-structured with structured tables; S-TEX: matching semi-structured with unstructured tables (containing two sub-datasets -- S-TEX-c for computer product data and S-TEX-w for watch product data); R-TEX: matching structured tables with unstructured data. Additionally, we utilize a geospatial database from OSM-FSQ-Pittsburgh~\cite{a23} with heterogeneous attributes and incorporate it to evaluate our system (denoted G-HET). For the training rate, i.e.,  $\%$ rate, we follow the setup in~\cite{a11} with the same splitting on training, validation and testing examples. The dataset statistics are summarized in Table \ref{tab:dat_stats} while experiments are conducted for both low- and sufficient-resource settings for the Machamp dataset including the geospatial database.  

\subsection{Baselines} We consider baselines from traditional deep learning techniques such as RNNs all the way to PrLMs based on full-scale PrLM fine-tuning and prompt-tuning. Below is a brief outline of the adopted baselines.

\noindent \textbf{DeepMatcher:} DeepMatcher~\cite{deepmatcher} is an RNN-based system for EM that proposes different modules, enabling diverse representation power to the learned embeddings. The best-performing module is then assessed for GEM.

\noindent \textbf{TDMatch:} TDMatch~\cite{a20} is an unsupervised library for EM utilizing graph creation and random walk generation. A supervised counterpart is also included as a baseline that employs an MLP classifier on top of the embeddings. The approach is dubbed TDMatch$^*$.  

\noindent \textbf{SentenceBERT:} SentenceBERT\cite{a21} leverages twin and triplet network architectures proposed to derive semantic representations which can be compared using the cosine similarity. These concepts can be extended to EM or GEM.

\noindent \textbf{BERT:} BERT~\cite{a22}, like the previous method, is a PrLM based on Transformers. Tuples in tables are processed as sentence pairs for the classification problem. 

\noindent \textbf{Ditto:} Ditto \cite{a5} is a system providing SoTA results on traditional EM benchmarks. Ditto can initialize with any PrLM such as BERT, RoBERTa, distilBERT, etc.

\noindent \textbf{DADER:} DADER~\cite{a9} is an EM system leveraging PrLMs and domain adaptation strategies. 

\noindent \textbf{Rotom:} Rotom~\cite{a8} also leverages PrLMs. It can use different data augmentation strategies and learns a policy to combine different data augmentation operators.

\noindent \textbf{PromptEM:}  PromptEM~\cite{a11} combines prompt-tuning and PrLMs to develop a system for GEM. By using domain-engineered templates, the backbone of a PrLM can be queried to provide a label word that maximizes class membership, i.e., `yes' or `no' to signify a binary relation.

\subsection{Implementation details:} Our experiments are conducted on an Ubuntu workstation equipped with an Nvidia RTX A6000 graphics card encompassing 48GiB of GPU memory. The PrLM is implemented in PyTorch version 1.12.1 paired with CUDA 11.3 drivers on a Python 3.9.7 environment. We use BERT-base~\cite{a22} as the backbone of our system which has already been implemented in the framework of adapter-transformers~\cite{pfeiffer2020AdapterHub}. To optimize memory usage mixed precision, i.e., fp16 is utilized for training the PrLM. The sequence length of the PrLM is set to 512, i.e., the maximum value possible. For all the task adapters, we sweep through learning rates lr =\{ 1e-4, 2e-4, 3e-4\} via parameter grid search using a batch size of 32. This is because adapter training requires slightly larger steps by default. For MLM training, we use lr=2e-4 with a batch size of 16 and 3 training epochs to avoid overfitting. AdamW is used as the optimizer and the fine-tuning epochs are chosen from {20, 30, 40, 50} depending on the dataset. We also optionally use TF-IDF-based text summarization introduced in Ditto~\cite{a5} during training. For evaluation, we chose the highest F1 on the validation set and report the results on the holdout test set.

\section{Results}
\label{section: Results}
We evaluate the proposed approach on the Machamp benchmark for both low and sufficient-resource. Our results are recorded in Table~\ref{table:hr} respectively. The observations and conclusions drawn from these results are discussed in the following subsections.

\subsection{AdapterEM vs Baselines}

\subsubsection{Low-Resource:}
Empirically, we observe that in 7 out of 8 tasks, AdapterEM performs better than full-scale PrLM fine-tuning. Stacking language adapters with task adapters has also shown some advantages over using a single task adapter across the board (e.g., S-HOM, S-HET, S-TEX-c, S-TEX-w, and G-HET).  The performance gains are more pronounced on type datasets for which we suspect catastrophic forgetting effects in  terms of the full-scale PrLM fine-tuning baselines. For example, on S-HET, we increase the margin by +34.4\% when compared to full-scale PrLM fine-tuning (although the best performance is recorded from the unsupervised TDmatch). On S-TEX-c we observe similar phenomena where AdapterEM performs better by a considerable margin of +10.8\% over full-scale PrLM fine-tuning. On other benchmarks, the results improve by a small margin with the exception of S-HOM where full-scale PrLM fine-tuning maintains 0.2\% F1 points over  AdapterEM.  On the other hand, PromptEM has achieved the bulk of the best performance in low-resource settings. However, AdapterEM is not far off from the estimates of prompt-tuning in 5 out of 8 scenarios. An analysis from the perspective of computational efficiency is presented in Section~\ref{sec:efficiency}.

\begin{table*}[!ht]
\setlength\tabcolsep{3pt}
\setlength\thickmuskip{0mu}
\setlength\medmuskip{0mu}
\caption{Performance$^\dag$ under different resource settings. Baseline results are obtained from \cite{a12, a3}.} 
\let\centering\relax  
\scalebox{1.1}{
\begin{tabular}{c  c c c c c c c c} 
\hline 
\rowcolor{gray!50}
\multicolumn{9}{c}{ \textbf{\textit{Results under low-resource settings }}}\\
\hline
Methods (F1 score) & R-HET & S-HOM & S-HET& S-REL & S-TEX-c& S-TEX-w & R-TEX & G-HET  \\ [0.5ex] 

\hline 

DeepMatcher  &   0.0 &   73.8 &  35.1 &  54.7 &  29.1 &  3.5 & 23.4 & 43.8\\ 

BERT        &  95.2 &  91.6 &  27.6 &  93.3 &  51.6 &  20.2 &  26.9 &  79.4\\ 
SentenceBERT  &  95.2 & 93.1 & 23.7&  90.8 &  55.3 &  23.5 & 42.9&  73.3 \\ 


\hline 
Ditto       &  92.7 &  90.2 &  24.5&  88.3 &  51.5 &  30.3 &  41.1&  80.5 \\ 
DADER       &  81.8 &  86.2& 54.6 & 91.7  &  25.6 &  20.5 &  37.2 &  68.4\\ 
Rotom      &  87.2 &  91.7&  26.5 & 91.7 & 60.5&  38.3&  48.5 & 77.6\\ 
\hline

TDmatch & 72.1 & 58.0  & \textbf{92.4}  & 91.3 &  18.0  & 21.3 &  59.5  & 72.9\\ %
TDmatch* & 6.3& 83.6 & 25.1 & 71.4 & 35.4 & 26.9 & 44.1 & 51.0\\ %
\hline
PromptEM\textsubscript{best} &  \textbf{100}  & \textbf{94.2} &  71.6 &  \textbf{95.0} & \textbf{73.6} &  \textbf{41.1} &  \textbf{61.4} &  \textbf{86.5}\\ %
\hline

AdapterEM (w/ task) &  \textbf{100} &  91.1&   73.8 &  \underline{94.2} &  61.4 &  24.4 &   39.6 &  80.1 \\ %
w/ SNLI & \textbf{100}  &  92.0&   51.5 &  89.0 &  32.7 &   22.7 &   53.3 &  \underline{84.1} \\ %

w/ TAPT\textsubscript{20} & \textbf{100} &  92.2 &  \underline{89.3}&  90.2&  67.2& \underline{40.1}& \underline{54.1} & 82.0\\
w/ TAPT\textsubscript{40} &  \textbf{100}  &  \underline{92.9} &  64.7 & 91.0 & \underline{71.3} &  23.2 &  50.6 &  76.7  \\ %

\rowcolor{gray!50}
\multicolumn{9}{c}{ \textbf{ \textit{Results under sufficient-resource settings } }}\\
\hline

            
DeepMatcher  &  93.6 &  86.1 &  29.1 & 56.7&  44.2 &  42.7 &  53.4 &  86.6 \\ 

BERT        & 95.5 &  93.8 &  46.0 &  90.5&  88.6&  72.8 & 63.1 &  90.4\\ 
SetenceBERT  &  69.6 &  87.4 & 69.7&  59.0&  79.8 &  50.2&  32.9 &  89.8  \\ 

\hline 

Ditto       &  \textbf{100}  &  93.1 &  61.6&  91.1&  81.8&  64.9&  62.7 &  90.1 \\ 
DADER       &  88.9&  87.4 &  57.1&  92.0& 24.4&  20.5 & 37.2 &  80.2\\ 

Rotom      &  \textbf{100}  &  94.7&  37.6&  94.4 &  90.5 &  73.9 &  65.7&  90.0\\ 


\hline
TDmatch &  72.1&  58.0 &  92.4&  91.3 &  18.0 &  21.3 &  59.5&  72.9\\ %
TDmatch* & 50.0 & 88.8 & 48.4 & 86.8 &75.1 & 65.8 & 56.4 & 75.5\\ %

\hline
PromptEM\textsubscript{best} & \textbf{100} &  \textbf{96.3} &  \textbf{93.0} &  96.2 &  92.1 &  74.9 &  66.4 &  \textbf{92.6}\\ %
\hline

AdapterEM (w/ task) &  \textbf{100}  & 95.3 & \underline{81.5} &  94.7 &  89.7&  68.6 &  68.0 &   \underline{92.4}\\ %
w/ SNLI & \textbf{100}  &  95.4 & 74.8  &  \textbf{97.6} &  89.3  &  64.4 &   65.0 &  90.4 \\ %

w/ TAPT\textsubscript{20} & \textbf{100} & \underline{95.9} & 78.6 &  88.5& 92.1 & \textbf{76.5} & 67.9& 91.2 \\
w/ TAPT\textsubscript{40} &\textbf{100}   & 95.5  & 78.9 &  90.9  &  \textbf{92.5} &  73.6 &  \textbf{68.1}  &  91.9\\ 

\hline 

    \multicolumn{9}{c}{
    \begin{minipage}{10cm}
      \small $^\dag$ Performance of all models on the Machamp benchmark. For each method, we report the F1 score. The best result is indicated with bold fonts in each block for low- and sufficient resource settings. The presence of an underscore indicates the best module in AdapterEM. Above, AdapterEM can utilize one of three kinds of adapters; task-only, SNLI pre-trained + task, or MLM-trained + task adapters. 
     \end{minipage}%
}\\

\end{tabular}}
    
\hfill\
\label{table:hr} 
\end{table*}

\subsubsection{Sufficient-Resource:}
All the available training data has been used for sufficient-resource settings to train each approach. In these settings, it is expected that adapter-based tuning will perform on par or better than full-scale PrLM fine-tuning.  Experimental results indicate that the adapter-based approach (i.e., AdapterEM) consistently outperforms full-scale PrLM fine-tuning on 7 of the 8 tasks with the exception of R-HET where we obtain an F1 score of 100\% similar to full-scale PrLM fine-tuning methods, i.e., Ditto and Rotom. Again we observe that the best-performing solution comes from stacking a language adapter with a randomly initialized or blank task adapter. Based on this evidence, language adapters inject some knowledge beneficial to downstream tasks.  Compared to the prompt-tuning-based method - PromptEM, we either perform almost on par or better than prompt-tuning. For example, on 4 tasks we outperform PromptEM (i.e., S-REL, S-TEX-w, S-TEX-c, and R-TEX) with a margin ranging between 0.4\% - 1.7\% respectively. On the remaining benchmarks, we perform nearly on par with PromptEM with the exception of S-HET (we record a margin loss of -11.5\% albeit better than full-scale PrLM fine-tuning). 
\begin{figure}[h!]
    \centering
    \includegraphics[width=8.6cm]{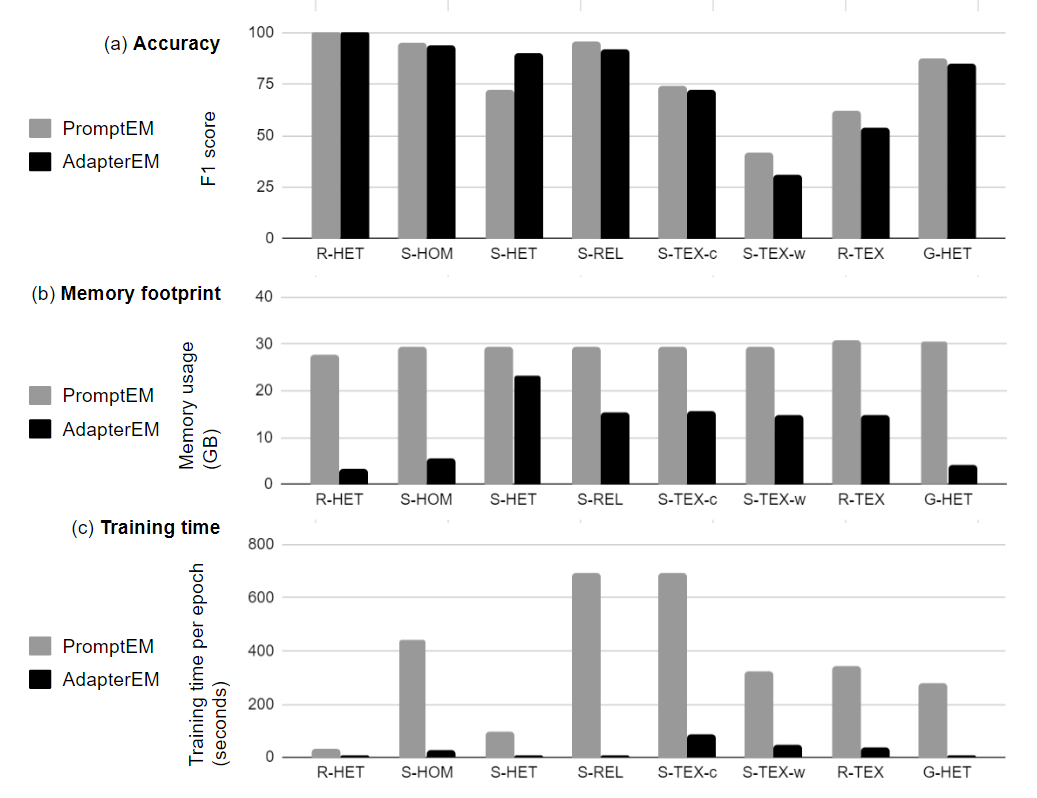}
    \caption{Computational footprint for low-resource settings.}
    \label{fig: cf}
\end{figure}
\subsection{Computational footprint}
\label{sec:efficiency}

To investigate our hypothesis that adapter-tuning can provide a computationally efficient alternative to existing approaches, we compare the computational footprint of our proposed AdapterEM approach with full-scale PrLM fine-tuning baselines, focusing on PromptEM due to its near-perfect dominance in terms of accuracy over other baselines (the one exception being TDMatch, which performs best for the S-HET matching task but requires an order of magnitude more execution time and memory usage, for example up to 120 hours and 130GB to achieve this, as also reported in~\cite{a12}). In order to make an overall comparison between PromptEM and AdapterEM from both an accuracy and computational footprint perspective, Figure \ref{fig: cf} compares F1-based accuracy (a), memory footprint (b), and training times (c), from which it can be observed that fine-tuning AdapterEM exhibits computational efficiency in terms of memory usage and training time. Although omitted in this study, we can extrapolate that this should hold true for sufficient-resource settings also, as more data is being used to train the models. This emphasizes the practicality of our adapter-based method for real-world deployment. Significantly, PromptEM and the baseline solutions examined in Section~\ref{section: Experiments} require $\approx$ 30G of GPU memory, whereas AdapterEM halves this requirement for many of the matching tasks (S-REL, S-TEX-c, S-TEX-w, R-TEX) and requires only around 10\% of the memory used by PromptEM for the R-HET and G-HET tasks. Since adapters typically involve fewer parameters, training steps can proceed faster and this property is accentuated by competitive training times, as can be observed in Figure \ref{fig: cf}(c), which compares the training times per epoch for AdapterEM and PromptEM, showing that AdapterEM requires only a small fraction of the PromptEM training times. The results show that the AdapterEM approach can provide a significant reduction in the overall computational footprint of EM tasks, which is critical given the ubiquity of applications requiring EM in domains ranging from E-commerce to science, among others.

\subsection{Discussion}
\label{section: Discussion}
Results presented in Table \ref{table:hr} show that AdapterEM performs better than full-scale PrLM fine-tuning in low-resource scenarios and is not far from the overall performance of PromptEM. In other words, AdapterEM can degrade the performance by a small margin in the low-resource setting compared to PromptEM. We find that this data-scarce setting combined with data imbalance presents a challenging scenario for AdapterEM. However, when using sufficient-resource settings, AdapterEM can yield comparable or even better performance than PromptEM. We also find that learning token-level language representations via the MLM objective can benefit both resource settings. Furthermore, AdapterEM tuning outperforms full-scale PrLM fine-tuning on sufficient-resource settings for GEM while avoiding a costly parameter update.  

\section{Conclusion}
\label{section: Conclusion}
We present a novel system called AdapterEM that leverages adapter-tuning as the underlying mechanism for fine-tuning a PrLM. By utilizing its modules, it can instantiate with randomly initialized task adapters whose weights are learned at the fine-tuning time or by combining the former with pre-trained adapters exploiting transferable features learned from auxiliary tasks such as SLNI or MLM training on target domain data. Our results demonstrate that our adapter-based approach can mitigate the effects of catastrophic forgetting on the Machamp EM benchmark. Moreover, AdapterEM reduces the computational complexity associated with using PrLMs, making it highly computationally efficient. In future works, we plan to investigate whether this approach can benefit from data augmentation techniques to improve performance. We also plan to investigate newer architectures of adapters as they evolve and to study their wide-ranging impact on EM.

\begin{acks}

A part of this paper is based on results obtained as part of a project, JPNP20006, commissioned by the New Energy and Industrial Technology Development Organization (NEDO) as well as JSPS KAKENHI Grant Number JP22H03694 and JST CREST Grant Number JP-MJCR22M2.

\end{acks}


\bibliographystyle{ACM-Reference-Format}
\bibliography{paperbib} 

\end{document}